\documentclass[journal]{IEEEtran}

\usepackage{amsmath}
\usepackage{amssymb}
\usepackage{array}
\usepackage{booktabs}
\usepackage{cite}
\usepackage{color}
\usepackage{courier}
\usepackage{graphicx}
\usepackage{helvet}
\usepackage{multirow}
\usepackage{subfigure}
\usepackage{tablefootnote}
\usepackage{times}
\usepackage{verbatim}

\newcommand\etal{{et~al.}}
\newcommand\ie{{i.e.}}

\newcommand\eg{{e.g.}}
\newcommand\fig{{Fig.}}
\newcommand\system{{ATTR}~}

\ifCLASSINFOpdf
\else
\fi

\begin{document}
\title{Aggregated Text Transformer for \\Scene Text Detection}

\author{Zhao Zhou, Xiangcheng Du, Yingbin Zheng, Cheng Jin
\thanks{The authors are with School of Computer Science, Fudan University, Shanghai, China. Zhao Zhou and Yingbin Zheng are also with Videt Lab, Shanghai, China. Zhao Zhou and Xiangcheng Du contributed equally to this work. Corresponding author: Cheng Jin.}
}

\maketitle
\begin{abstract}
  This paper explores the multi-scale aggregation strategy for scene text detection in natural images. We present the \emph{Aggregated Text TRansformer} (ATTR), which is designed to represent texts in scene images with a multi-scale self-attention mechanism. Starting from the image pyramid with multiple resolutions, the features are first extracted at different scales with shared weight and then fed into an encoder-decoder architecture of Transformer. The multi-scale image representations are robust and contain rich information on text contents of various sizes. The text transformer aggregates these features to learn the interaction across different scales and improve text representation. The proposed method detects scene texts by representing each text instance as an individual binary mask, which is tolerant of curve texts and regions with dense instances. Extensive experiments on public scene text detection datasets demonstrate the effectiveness of the proposed framework.    
\end{abstract}


\IEEEpeerreviewmaketitle

\section{Introduction}
\label{sec:intro}

Scene text detection consists of finding in a natural image the text regions that are possibly viewed from different viewpoints and in the presence of clutter and occlusion. Typical applications include the text information processing tasks, such as scene understanding and photo translation. 

\begin{figure}[t]
    \centering
    \includegraphics[width=.9\linewidth]{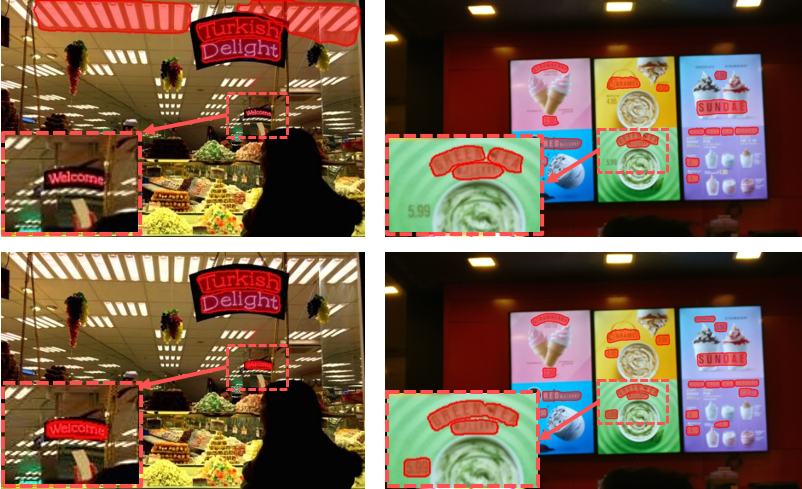}
    \caption{Scene text detection results with single-scale input (top) and with multi-scale aggregation (bottom). The single-scale setting may assign low confidence for the scene texts with small size.}
    \label{fig:motivation}
\end{figure}

Recent years have seen great progress in the area of scene text detection~\cite{liao2017textboxes,deng2018pixellink,ma2018arbitrary,wang2019shape,liao2020real,zhang2020deep,he2021most}. Various schemes based on the deep neural network are introduced and achieve remarkable results. The scene text detection problem, however, still remains challenging due to the irregular shapes and distinction of scene text, as well as diverse scales. Recent methods usually learn text representations by feeding the input image through a deep network. One potential problem for the single-scale setting is it tends to assign low confidence for the scene texts with different region sizes, especially for the small ones. Some examples are illustrated in the top row of \fig~\ref{fig:motivation}, where the inference is with the single-scale manner; adding multi-scale aggregation can reduce the number of missing text instances.

\begin{figure*}[t]
    \centering
    \includegraphics[width=.75\linewidth]{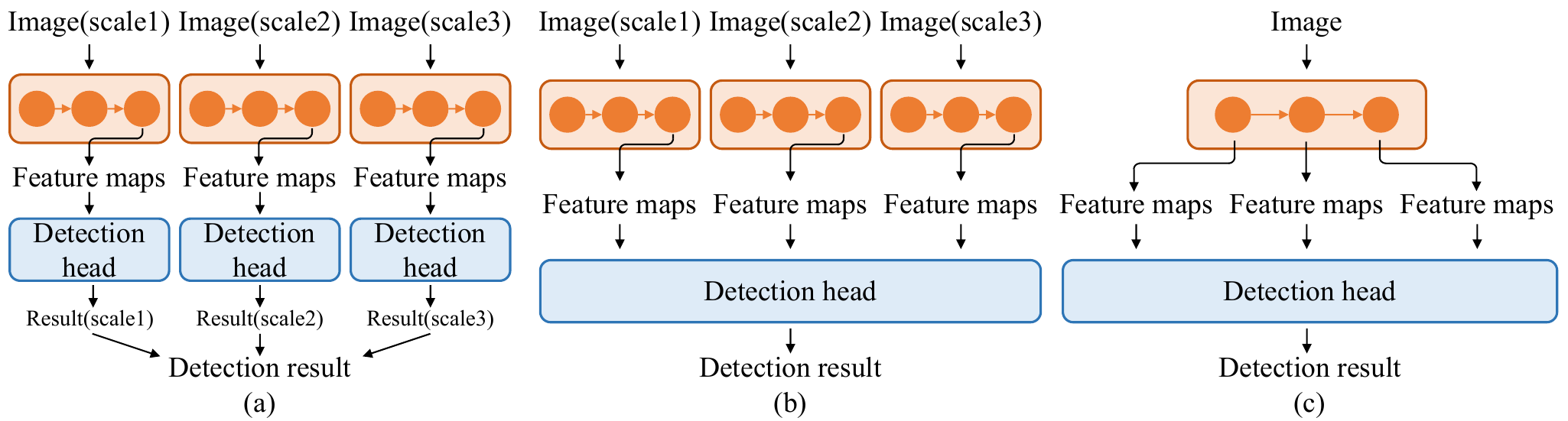}
    \caption{Overview of multi-scale aggregation strategies. (a) Initial results are computed separately from each image scale and the detection result is fused with a strategy like NMS. (b) Feature maps are obtained from a multi-scale image pyramid and fed into one detection head. (c) Aggregation of multi-scale feature maps for the detection results.}
    \label{fig:multiscalepipeline}
\end{figure*}

In this work, we explore the multi-scale aggregation strategy for text detection of complex natural scene images. Multi-scale approaches can provide a comprehensive representation of an image with a better sense of the context at multiple resolutions. Existing scene text detection methods are usually with three types of multi-scale settings. 
The first strategy is to train the network in a fixed input size and test the images on multiple scales followed by a late fusion~\cite{zhang2019look,zhu2021textmountain} (\fig~\ref{fig:multiscalepipeline}(a)), and the second is to extract feature maps from the image pyramid with different scales and then sent to the detection head~\cite{xue2019msr} (\fig~\ref{fig:multiscalepipeline}(b)). There is also the multi-scale feature aggregation (\fig~\ref{fig:multiscalepipeline}(c)), by learning multi-scale feature maps from a unique image by the feature pyramid network~\cite{zhou2017east,liao2020real,tang2022few}.
In this paper, we are inspired by \cite{lowe2004distinctive} that learns the keypoints from levels of the image pyramid and follow the pipeline that takes feature maps from the image pyramid as the input of the detection head. In this manner, the detection framework is adaptive to various scales and aspect ratio texts, leading to better scene text detection results. 

Particularly, we employ a Transformer architecture in our framework. The construction of robust representations is of fundamental importance for a scene text detector. It requires learning features that are discriminative enough to understand text regions with highly various shapes and fonts but also robust enough to clutters or extreme viewpoints. In the last few years, Vision Transformers have emerged as powerful representations for various image understanding tasks such as image classification~\cite{dosovitskiy2020image}, object detection~\cite{carion2020end}, and semantic segmentation~\cite{xie2021segformer,cheng2022masked}. In this paper, we present the \emph{Aggregated Text TRansformer} (ATTR), which follows the encoder-decoder architecture and is designed to utilize the self-attention mechanism across the image hierarchy and generate better features for text images.

Our framework also designs a novel text representation for arbitrary scene texts. Previous arbitrary shape text representations can be classified into two types: segmentation-based and regression-based. Segmentation-based methods~\cite{liao2020real,liao2022real,baek2019character,wang2019shape} represent text regions with pixel-level classification masks that are fitting of arbitrary shapes of texts accurately, but they require additional prediction and complex post-processing to distinguish between adjacent text areas. Regression-based methods~\cite{liu2020abcnet,tang2022few} regress the text boundaries directly without any post process, but in the complex case, the bounds of the regression cannot be fitted precisely to the textual region.
In this paper, we consider each text instance as a set of pairs, that consist of a text-background prediction and a text query vector. The text query vector is used to get the binary mask prediction via a dot product with the text embedding obtained from the multi-scale transformer encoder. 
This text instance representation can not only get tight text boundaries with the arbitrary shape like previous segmentation-based methods but also be easily separated between neighboring text instances as the regression-based methods.
We summarize our main contributions as follows.
\begin{itemize}
    \item Our framework significantly enhances the baseline text detector for representing texts in scene images using the multi-scale image aggregation and the self-attention mechanism of the Transformer. The proposed \system achieves state-of-the-art scene text detection performance on several public datasets.
    \item In our framework, each text instance is represented by an individual binary mask during decoding, which can separate neighboring text instances better and get tighter text instance boundaries.
    \item Different from previous text detection methods that mainly consider late fusion or feature level aggregation, we explore aggregating multi-scale features from the same projection on the image pyramid as the inputs of a Transformer encoder, which reduces the inconsistency of features from different network layers.
\end{itemize}

\section{Related Work}
\label{sec:related}

\vspace{0.03in}
\noindent\textbf{Scene text detection} has been an active research topic in computer vision recently and many approaches based on deep neural networks have been investigated. For instance, Zhou \etal~\cite{zhou2017east} predicts text instances of arbitrary orientations and quadrilateral shapes in text images. The regression-based methods are limited to representing accurate bounding boxes for irregular shapes, and therefore the segmentation-based methods are introduced. Wang \etal~\cite{wang2019shape} therefore performs progressive scale expansion on segmentation maps, and Liao \etal~\cite{liao2020real} adaptively learns the thresholds for the binarization process. For the text instance representation, Deng \etal~\cite{deng2018pixellink} predicts text/non-text and link prediction on scene text images, and Liu \etal~\cite{liu2021abcnet} introduces a concise parametric representation of scene text with the Bezier curves. Multi-scale networks with share weights are also used in Xue \etal~\cite{xue2019msr} to enhance the shape regression of text regions. Our approach is fundamentally different from theirs in the design for multi-scale aggregation schemes and text instance representations. Moreover, as will be shown in the experiments, the proposed \system outperforms these previous methods.

\vspace{0.03in}
\noindent\textbf{Vision Transformer.}
Transformer structures have attracted increasing research interest in the field of computer vision and a comprehensive survey can be found in Han \etal~\cite{han2022survey}. Dosovitskiy \etal~\cite{dosovitskiy2020image} is the first to employ a transformer on sequences of flatten image patches and achieve state-of-the-art image classification results. Touvron \etal~\cite{touvron2021training} then introduces a data-efficient training procedure, and Liu \etal~\cite{liu2021swin} incorporates the hierarchical structure with shifted windows. Carion \etal~\cite{carion2020end} designs the DETR for end-to-end object detection, by treating the task as a set prediction problem. Zhu \etal~\cite{zhu2020deformable} further combines a sparse spatial sampling of deformable convolution and improves DETR. For the semantic segmentation task, Transformer-based approaches are also proposed, such as SegFormer~\cite{xie2021segformer} and Mask2Former~\cite{cheng2022masked}. Specifically for text detection task, Tang \etal~\cite{tang2022few} employs transformer-based architecture for scene text detection, and Zhang \etal~\cite{zhang2022text} designs a transformer-based text spotting framework which regresses control points coordinates.

\vspace{0.03in}
\noindent\textbf{Multi-scale approaches} have a long history in computer vision. 
Lowe \etal~\cite{lowe2004distinctive} introduces the SIFT descriptor extracted from octaves of image scale space. Spatial pyramid~\cite{lazebnik2006beyond,he2015spatial} is widely used in scene understanding and image recognition tasks, by computing features of each region on different grids and concatenating them to form a complete representation. 
The feature pyramid method can be employed to assemble feature maps at multiple scales~\cite{zhao2017pyramid}. 
Multi-scale feature aggregation is also used to enhance the Vision Transformer~\cite{fan2021multiscale}. Another work more related to ours is CrossViT~\cite{chen2021crossvit}, which applies a dual-branch vision transformer to extract multi-scale feature representations for image classification. Our work builds on the observation that image patches of different sizes produce stronger image features, incorporate the characteristics of scene images, and finally show the effectiveness of multi-scale aggregation for the scene text detection task.

\begin{figure}[t]
  \centering
  \includegraphics[width=.9\linewidth]{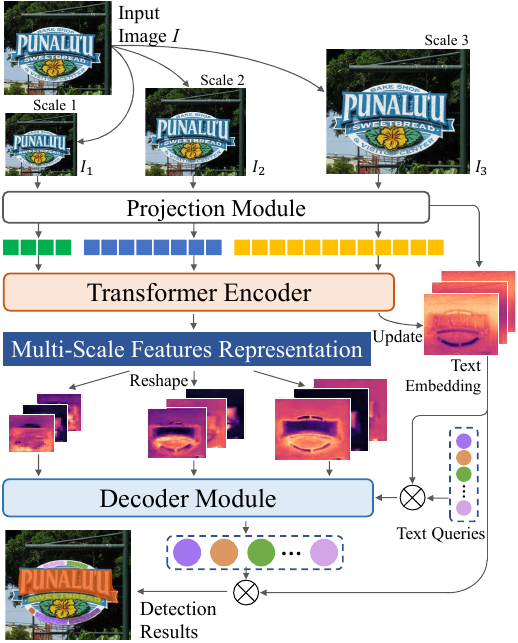}
  \caption{The proposed framework. Given a scene image, we augment it into a multi-scale image pyramid. The multi-scale image representations are computed with a projection module with shared weights for different scales and a Transformer encoder that learns multi-scale contexts. The text instances are then detected with the decoder module.}
  \label{fig:arch}
\end{figure}

\section{Framework}
\label{sec:Framework}
\fig~\ref{fig:arch} shows the pipeline of our framework. Given a scene image $I$, we augment it into a multi-scale image pyramid. Following settings of many previous works, three scales are used with scale factors of \{$\frac{1}{2}$,1,2\} and we denote these inputs as $\{I_1, I_2, I_3\}$. The multi-scale image representations are computed with a projection module with shared weights for different scales and a Transformer encoder that learns multi-scale contexts. Meanwhile, text embedding is initialized with the intermediate feature map from the last scale and updated with the multi-scale features. With multi-scale image representations and text instance representation from text embedding and text queries, text instances are obtained with the decoder module which consists of several scale-wise decoders.

In the following parts of this section, we first introduce the proposed multi-scale image representations, including the projection of the multi-scale image pyramid and the Transformer encoder design. Then, we will elaborate on the details of the text instance representation, decoders, and model training of our proposed method.

\subsection{Multi-Scale Image Representations}

The construction of robust image representations is important for vision tasks. Many previous scene text detectors employ the FPN~\cite{lin2017feature} based feature extractor. Instead of this type of multi-scale feature fusion, we operate on the multi-scale image pyramid, and the representation for each scale is directly learned from the last layer of the projection module. We are motivated by the success of ViT~\cite{dosovitskiy2020image} and CrossViT~\cite{chen2021crossvit}, and we believe that learning multi-scale self-attention from homogeneous feature maps is more feasible for the self-attention mechanism.

\begin{figure}[t]
  \centering
  \includegraphics[width=\linewidth]{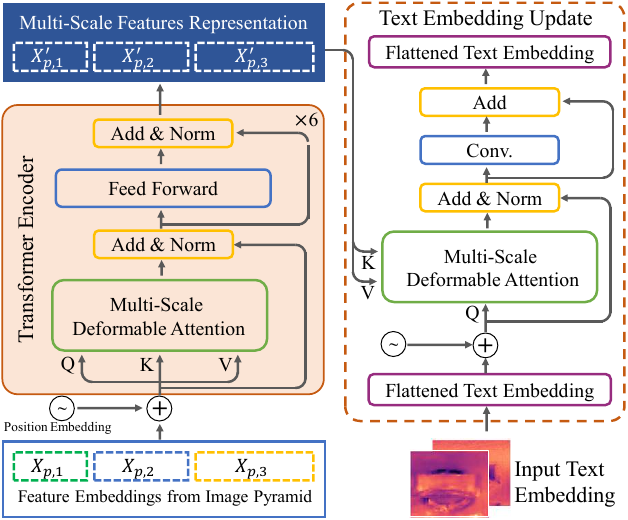}
  \caption{Taking feature embeddings from image as input, we first employ the Transformer Encoder to compute the multi-scale image representations. The text embedding is updated with the image representations as input.}
  \label{fig:encoder}
\end{figure}

\vspace{0.03in}
\noindent\textbf{Projection module.} 
One intentionally simple setting for the projection module is the linear projection, as in~\cite{dosovitskiy2020image}. We reshape the image $I\in \mathbb{R}^{H\times W \times 3}$ into a sequence of flatten patches $I_p^1,I_p^2, ..., I_p^{n_0}\in \mathbb{R}^{d_0}$, where $n_0=\frac{HW}{P^2}$, $d_0=3P^2$, and $P$ is the patch size. With a trainable linear projection function $\phi_{\textrm{lp}}$, $I_p$ is then map to the $c$ dimension feature embedding $X_p=[x_p^1,x_p^2,...,x_p^{n_0}]$ where $x_p^i=\phi_{\textrm{lp}}(I_p^i)\in \mathbb{R}^c$. For the image pyramid $\{I_1,I_2,I_3\}$, we can obtain \{$X_{p,1}$,$X_{p,2}$,$X_{p,3}$\}. The multi-scale feature embeddings are the concatenation of each scale.

One drawback of linear projection for our framework is its representation ability, especially for the text regions that are usually small and irregular. Therefore, some alternative designs are employed for our projection module. We first try a very simple convolution network $\phi_{\textrm{conv}}$ with one convolutional layer followed by three convolutional blocks\footnote{Both the  layer and blocks are with a stride of 2.}.
The feature embeddings of $\phi_{\textrm{conv}}$ keep the same dimension and patch number as that of $\phi_{\textrm{lp}}$. We also have the feature embedding $\phi_{\textrm{res}}$ by replacing the convolutional blocks with the residual blocks \cite{he2016deep}. The effect of different projection modules will be presented in the ablation study. It seems $\phi_{\textrm{res}}$ makes a good trade-off between representation ability and efficiency.

\begin{figure}[t]
  \centering
  \includegraphics[width=\linewidth]{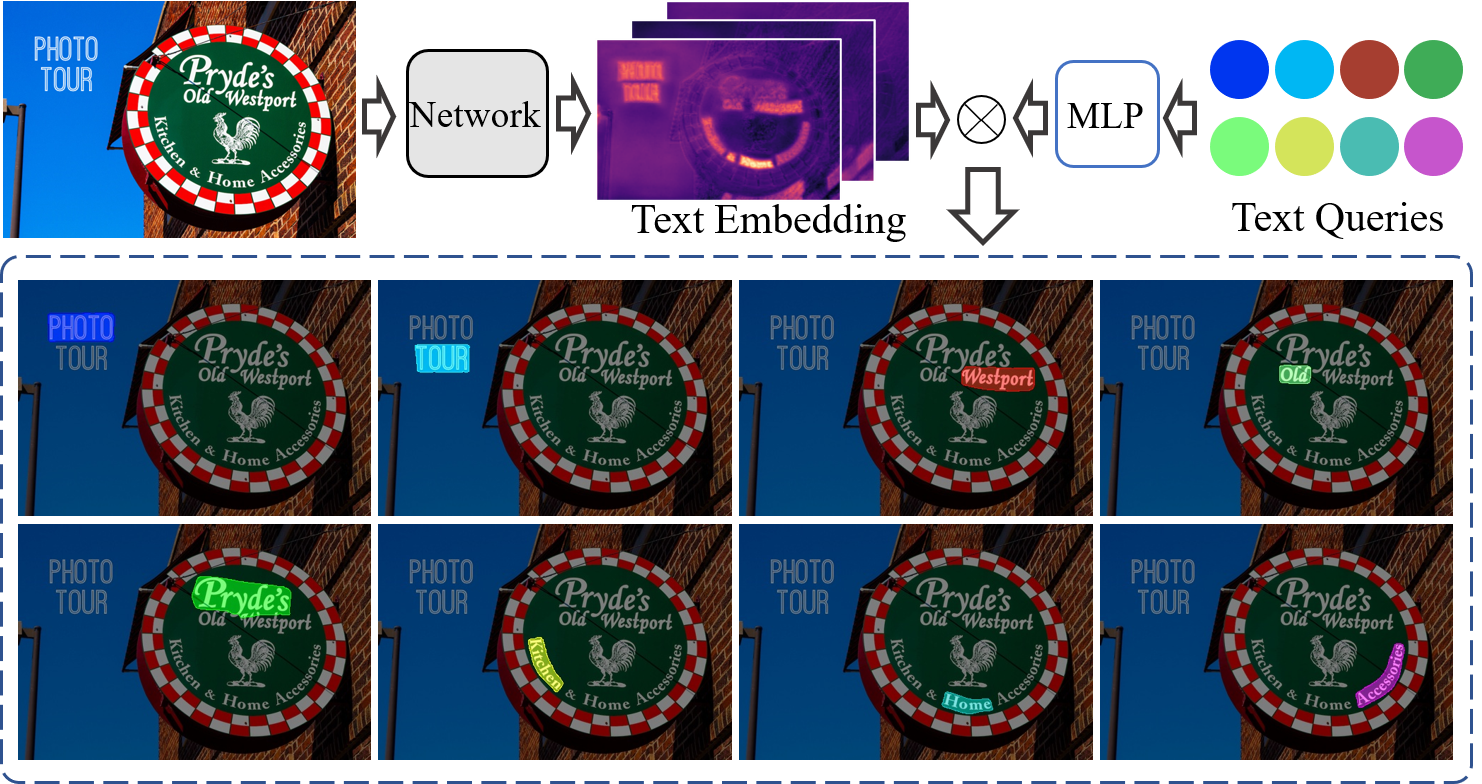}
  \caption{Text instance representation generation and the visualization of binary masks. Each result image indicates one mask (text instance) in the original image.}
  \label{fig:text_represent}
\end{figure}

\vspace{0.03in}
\noindent\textbf{Transformer encoder.} 
The detailed structure of the Transformer encoder is on the left of \fig~\ref{fig:encoder}. The training of vision Transformer based models usually requires a large amount of image data~\cite{dosovitskiy2020image,carion2020end}. However, the image number of the text detection training set is usually limited, and the synthetic images lack background changes. Here we employ the deformable attention~\cite{zhu2020deformable}, which is robust on mid-size data and reduces the complexity. Each encoder unit contains multi-scale deformable attention and a feed-forward net. Different from~\cite{zhu2020deformable} that considers feature maps from multiple layers, our input $Z_{k}$ is based on the representation from the image pyramid and the position embeddings, \ie, $Z_k=X_{p,k}+E_{pos}$. The final image representations \{$X'_{p,1}$,$X'_{p,2}$,$X'_{p,3}$\} are obtained with six encoder units.

\vspace{0.03in}
\noindent\textbf{Text embedding} aims to present potential text regions and to further generate the binary text mask prediction. During our attempt at the projection module, we observe that part of the feature maps from early layers contain text-like edges. This is consistent with the observations from previous research on image recognition~\cite{zeiler2014visualizing}. We use these feature maps as the initial text embedding. However, not all of them represent text regions, and thus we update the text embedding with the help of multi-scale image representations. As shown on the right of \fig~\ref{fig:encoder}, another multi-scale deformable attention is utilized to update the text embedding. 

\subsection{Detecting Text Instances}
The text instances are detected based on the multi-scale image representations, decoder module, and text instance representation which will be first introduced in this part.

\begin{figure}[t]
  \centering
  \includegraphics[width=.85\linewidth]{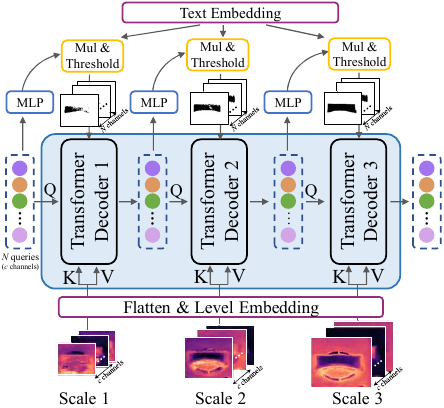}
  \caption{Structure of scale-wise decoders. Due to space limitation, we only visualize 3 transformer decoders; for the settings of 6/9 decoders, the image representations from scale 1 to 3 will be repeated as the input in a round robin fashion.}
  \label{fig:decoder}
\end{figure}

\vspace{0.03in}
\noindent\textbf{Text instance representation.} Different from previous text detection methods, we employ a novel text representation method inspired by recent instance segmentation~\cite{wang2020solov2}, by predicting a set of binary masks for each text instance. The pipeline is illustrated in \fig~\ref{fig:text_represent}. Given the $t$-th query $Q_t\in \mathbb{R}^{1\times 1\times c}$ and the text embedding $E\in \mathbb{R}^{h\times w\times c}$, we compute the binary mask $M_t$ value at location ($i$,$j$) as
$$
  M_{t}(i, j) = \text{sigmoid}(E(i,j)\otimes \mathrm{MLP}(Q_t)).
$$
Here MLP($\cdot$) transfers $Q_t$  with the same dimension and $\otimes$ indicates the dot-product. 
There are two advantages to this set of text representations. First, previous segmentation-based approaches usually need extra components for the text instances, \eg, the threshold map in DB~\cite{liao2020real} and the centripetal shifts in CentripetalText~\cite{sheng2021centripetaltext}, ours simplify the design of text instance segments. Second, our approach is benefited from multi-scale image representations directly and can obtain a more accurate boundary of text instances with different sizes and shapes.
Some qualitative comparisons are illustrated in \fig~\ref{fig:text_represent_comp} and more comparisons with previous representations will be conducted in the experiments.

\vspace{0.03in}
\noindent\textbf{Decoder module.} 
\fig~\ref{fig:decoder} show the structure of the scale-wise decoders. We follow the design of \cite{cheng2022masked}. In each stage, the image representation from one scale is sent to the decoder. The text queries are also the input of the module with two branches. The first branch is directly fed into the decoder and the second follows the same steps as \fig~\ref{fig:text_represent} to generate text masks. The Transformer decoder contains a masked attention~\cite{cheng2022masked} followed by a self-attention and outputs the new queries. As the text embedding is fixed during the whole procedure, the text masks will be refined with the update of queries.

\subsection{Model Training}

\vspace{0.03in}
\noindent\textbf{Loss.}
To train our model, the loss function $L$ is expressed as a combination of mask loss and classification loss:
$$
L = L_{\textrm{mask}} + \lambda_{\textrm{cls}}L_{\textrm{cls}},
$$
where $L_{\textrm{cls}}$ is the loss for text/non-text prediction, and $L_{\textrm{mask}}$ is the loss for binary mask of text instances. We set $\lambda_{\text{cls}} = 0.4$ for predictions matched and $\lambda_{\text{cls}} = 0.02$ for predictions that have not been matched with any ground-truth.

$L_{\textrm{mask}}$ is the sum of the binary cross-entropy loss and the dice loss over each query at different sampled positions, \ie,
$$L_{\textrm{bce}} = \frac{1}{N}\sum_{t=1}^{N}\sum_{i=1}^{K}[y_t^i\log(\hat{y}_t^i)+(1-y_t^i)\log(1-\hat{y}_t^i)]$$
$$
L_{\textrm{dice}} = \frac{1}{N}\sum_{t=1}^{N}(1 - \sum_{i=1}^{K}\frac{2 \cdot \hat{y}_t^i \cdot y_t^i}{\hat{y}_t^i + y_t^i})
$$
where $\hat{y}_t^i \in [0,1]$ and $y_t^i \in \{0,1\}$ are value of predictions (with logits) and ground-truth of query $Q_t$ at the $i$-th sampled position, $N$ and $K$ are the number of the queries and the sampled positions.

For the text/non-text classification, we adopt the Cross-Entropy loss which can be formulated as:
$$
L_{\textrm{cls}} = \frac{1}{N} \sum_{t=1}^{N}[l_{t} \cdot \log(p_{t}) + (1-l_{t}) \cdot \log(1-p_{t})]
$$
where $l_{t}$ and $p_{t}$ represents the label and the predicted probability of query $Q_t$, respectively.

\noindent\textbf{Post-processing.}
The final text binary marks are computed based on text embedding and text queries, with the same steps as \fig~\ref{fig:text_represent}. 
There is a text/non-text prediction by a full-connected layer connected after the text query.
The confidence of each text instance is calculated by multiplying the class confidence by averaged foreground binary mask probability. 
\section{Experiments}
\label{sec:exp}
We examine the proposed approach on five popular text detection benchmarks: Total-Text~\cite{ch2017total}, CTW1500~\cite{liu2019curved}, ICDAR2019-ART~\cite{chng2019icdar2019}, ICDAR2015~\cite{karatzas2015icdar}, and MSRA-TD500~\cite{yao2012detecting}. 
{Among them, Total-Text, CTW1500, and ICDAR2019-ART aim to detect the curved texts, and ICDAR2015 and MSRA-TD500 are for detecting the multi-oriented texts.} 
The evaluation follows the standard protocols of these benchmarks. The Total-Text dataset consists of 1255 training images and 300 testing images. It contains not only horizontal and multi-oriented text instances but also curved texts. The images are annotated at the level of the word by polygons. The experiments on Total-Text are designed to exploit alternative settings of our framework. The CTW1500 dataset contains 1000 training images and 500 testing images. Each text instance annotation is a polygon with 14 vertexes to define the text region at the level of text lines. The text instances include both inclined texts as well as curved texts. ICDAR2019-ART is a large arbitrary shape scene text benchmark. It contains 5,603 training images and 4,563 testing images. ICDAR2015 contains 1000 training images and 500 testing images in English, most of which are severely distorted or blurred. All images are annotated with quadrilateral boxes at the word level. The MSRA-TD500 dataset consists of 300 training images and 200 testing images. It is a multilingual dataset focusing on oriented text lines, and large variations of text scale and orientation are presented. Following previous works \cite{tang2022few,liao2020real,long2018textsnake}, we include HUST-TR400 \cite{yao2014unified} as the extra training data in the fine-tuning stage. 

\begin{table}[t]
  \centering
  \footnotesize
  \caption{Multi-scale performance under different settings. ``P'', ``R'', and ``F'' represent {Precision}, {Recall}, and {F-measure}.}
  \label{tab:ms}
  \begin{tabular}{c|c|c|ccc|c}
  \hline
  \multicolumn{3}{c|}{Setting} & P & R & F & fps\\ 
  \hline
  \multicolumn{2}{c|}{} & $I_1$ & 90.6 & 82.4  & 86.3 & {15.4}\\ 
  \multicolumn{2}{c|}{(a) Single-scale} & $I_2$ & 89.4 & 85.1 & 87.2 & {11.7}\\ 
  \multicolumn{2}{c|}{} & $I_3$ & 89.8 & 85.7 & 87.7 & {5.8}\\
  \hline
  & &  $I_1$+$I_2$ & 90.0 & 86.6 & 87.8 & {6.7}\\
  (b)&Late fusion & $I_2$+$I_3$ & 89.1 & 86.3 & 87.7 & {4.0}\\
  Multi-& & $I_1$+$I_2$+$I_3$ & 88.7 & 87.0 & 87.9  & {3.1}\\
  \cline{2-7}
  scale&\multirow{3}{*}{\system} & {$I_1$+$I_2$} & {90.9} & {87.6} & {89.3} & {8.6}\\
  & & $I_2$+$I_3$ & 90.7 & 87.5 & 89.1 & {5.8}\\
  & & $I_1$+$I_2$+$I_3$ & 91.9 & 88.3 & 90.1 & {4.6}\\
  \hline
  \end{tabular}
\end{table}

\begin{table}[t]
  \centering
  \caption{{Comparison of aggregation approaches with different number of scales.}}
  \label{tab:aggregation}
  \footnotesize
  {
  \begin{tabular}{c|cccc|cccc}
      \hline
      {Aggregation} & \multicolumn{4}{c|}{{Two-scales}} & \multicolumn{4}{c}{{Three-scales}}\\ 
      {approach}  & {P} & {R} & {F} & {fps} & {P} & {R} & {F} & {fps} \\ 
      \hline
      (a) & 90.7 & 87.5 & 89.1 & {5.8} & {91.9} & {88.3} & {90.1} & {4.6} \\
      (b) & 90.6 & 84.1 & 87.2 & {7.6} & {88.6} & {86.3} & {87.4} & {6.5} \\
      \hline
  \end{tabular}
  }
  \\
  \vspace{0.05in}
  {\footnotesize{(a) Image pyramid level}}~~~~~~~~~~~~~~~~~~{\footnotesize{(b) Feature map level}}~~~~\\
  \includegraphics[width=.43\linewidth]{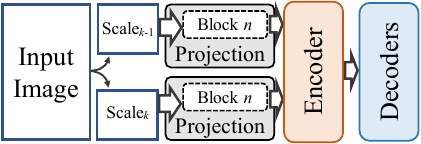}~~
  \includegraphics[width=.43\linewidth]{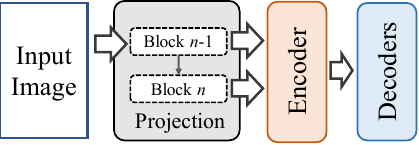}\\
\end{table}

\begin{table}[h]
  \centering
  \footnotesize
  \caption{Effect of the \system components: (a) projection module; (b) encoder module; (c) decoder module. {We report both performance and inference speed under different settings.}}
  \label{tab:components}
    \begin{tabular}{c|c|ccc|c}
      \hline
      \multicolumn{2}{c|}{Setting} & P & R & F & {fps}\\ 
      \hline
      \multirow{5}{*}{(a)}&Projection module: $\phi_{\textrm{lp}}$ & 83.4 & 79.5 & 81.4 & {5.1}\\
    &Projection module: $\phi_{\textrm{conv}}$ & 91.5 & 77.2 & 83.8 & {5.0}\\ 
    &{$\phi_{\textrm{res}}$ (2 residual blocks)} & 91.0 & 86.3 & 88.6 & {3.1}\\ 
    &{$\phi_{\textrm{res}}$ (3 residual blocks)} & 91.9 & 88.3 & 90.1 & {4.6}\\
      &$\phi_{\textrm{res}}$ (4 residual blocks) & 91.0 & 86.9 & 88.9 & {5.0}\\ 
      \hline
      \multirow{3}{*}{(b)}  &\system (conv. encoder) & 91.0 & 85.6 & 88.2 & {6.0}\\
      &\system (Transformer encoder) &91.9 & 88.3 & 90.1 & {4.6}\\
      &{\system (w/o text embedding update)} & {88.8} & {86.2} & {87.5} & {4.7}\\
        \hline
        \multirow{5}{*}{(c)}
        &\system (w/o decoder) & 80.8 & 69.9 & 75.0 & {5.6}\\
        &\system (3 decoders) & 91.8 & 86.4 & 89.0 & {5.0}\\
        &\system (6 decoders) & 91.8 & 86.9 & 89.3 & {4.7}\\
        &\system (9 decoders) &91.9 & 88.3 & 90.1 & {4.6}\\
        \hline
        \end{tabular}
\end{table}
\begin{figure}[t]
  \centering
  \includegraphics[width=\linewidth]{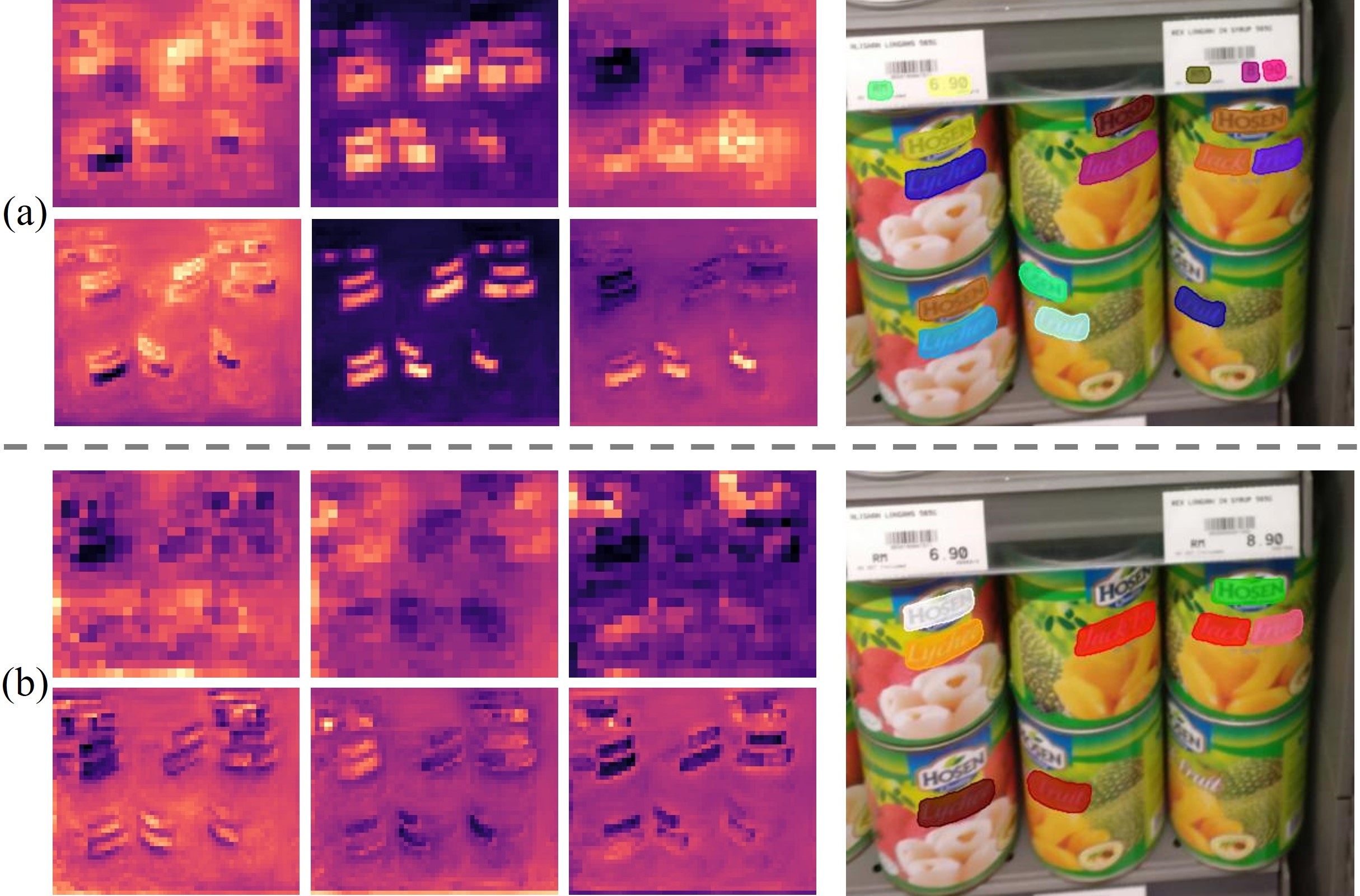}
  \caption{Visualization of feature maps and the detection results on a sample image from TotalText. The heatmaps correspond to the feature maps of the 1st to 3rd channels after encoders. The first row and the second row are from different scales. (a) the proposed ATTR. (b) feature-level scale aggregation.}
  \label{fig:single_multi_compare}
\end{figure}
\vspace{0.03in}
\noindent\textbf{Implementation details.}
We use an AdamW~\cite{loshchilov2018decoupled} optimizer and a step learning rate schedule. We use an initial learning rate of 0.0001 and a weight decay of 0.05 for all datasets. A learning rate multiplier of 0.1 is applied to the network, and decrement the learning rate by a factor of 10 at 0.9 and 0.95 of the total training steps. The data augmentation for the training data includes random scaling, random rotation, random cropping, large-scale jittering and random flipping. To obtain the model, the network is first initialized with pretraining on the SynthText 150K dataset~\cite{liu2020abcnet} with batch size of 16 and 40$k$ iterations, and then fine-tuned on each corresponding dataset for other 40$k$ iterations. We keep the aspect ratio of test images during inference and resize the shorter sides to 640. 
{We empirically set the number of text queries to 100.} 
Our default setting for \system is an image pyramid with three scales, three residual blocks for the projection module, and nine scale-wise decoders in the decoder module. 
{The inference speed is measured in the frames-per-second (fps) on a single NVIDIA TITAN Xp GPU with a batch size of 1 by taking the average runtime on the entire validation set including image loading, model forwarding, and post-processing time.}

\subsection{Ablation Study}
We first perform an ablation study on the Total-Text dataset, including the effect of multi-scale aggregation, different components in the proposed ATTR, and the model training.

\vspace{0.03in}
\noindent{\textbf{Multi-scale aggregation.}} 
{Recall that the proposed aggregated text Transformer is generated from three image scales, based on the projection module individually. In Table~\ref{tab:ms}(a) we list the detection results generated by performing the framework on a single image scale only. We can observe that a scale with higher resolution tends to improve the recall rate, while the precision is stable. The inference speed also depends on the scale, as large input leads to more tokens and increases the computational costs.
Table~\ref{tab:ms}(b) shows the performances of both our approach and the late fusion from multi-scale inputs (\fig~\ref{fig:multiscalepipeline}(a)). Generally speaking, the late fusion strategy improves recall and keeps the detecting precision. The proposed \system with three-scale aggregation further boosts both metrics and leads to a gain of 2.2 for the F-measure, indicating that ours is better for scale aggregation.}

We also conduct a comparison of the proposed method with feature map level scale aggregation (\fig~\ref{fig:multiscalepipeline}(c)). To make a fair comparison, we use input images with the same resolution and the same encoder-decoder structure. 
Both two-scale and three-scale aggregations are evaluated. 
Table~\ref{tab:aggregation}(a) and Table~\ref{tab:aggregation}(b) illustrate the pipeline of two-scale aggregation. For feature map level aggregation, the feature maps from the last two blocks have the same number and shape of feature embeddings that will be sent to the Transformer encoder, to ensure the identical token length and dimension.
As shown in the table, using our aggregation strategy leads to the recall of 87.5 (two-scale) and 88.3 (three-scale), which is significantly higher than feature-level scale aggregation. We also observe that the two-scale image pyramid level aggregation outperforms the three-scale feature map level aggregation.
The visualization of corresponding aggregation methods is illustrated in \fig~\ref{fig:single_multi_compare}, where we can see that our strategy can capture more accurate text boundaries while reducing background interference.

\begin{figure}[t]
  \centering
  \includegraphics[width=.85\linewidth]{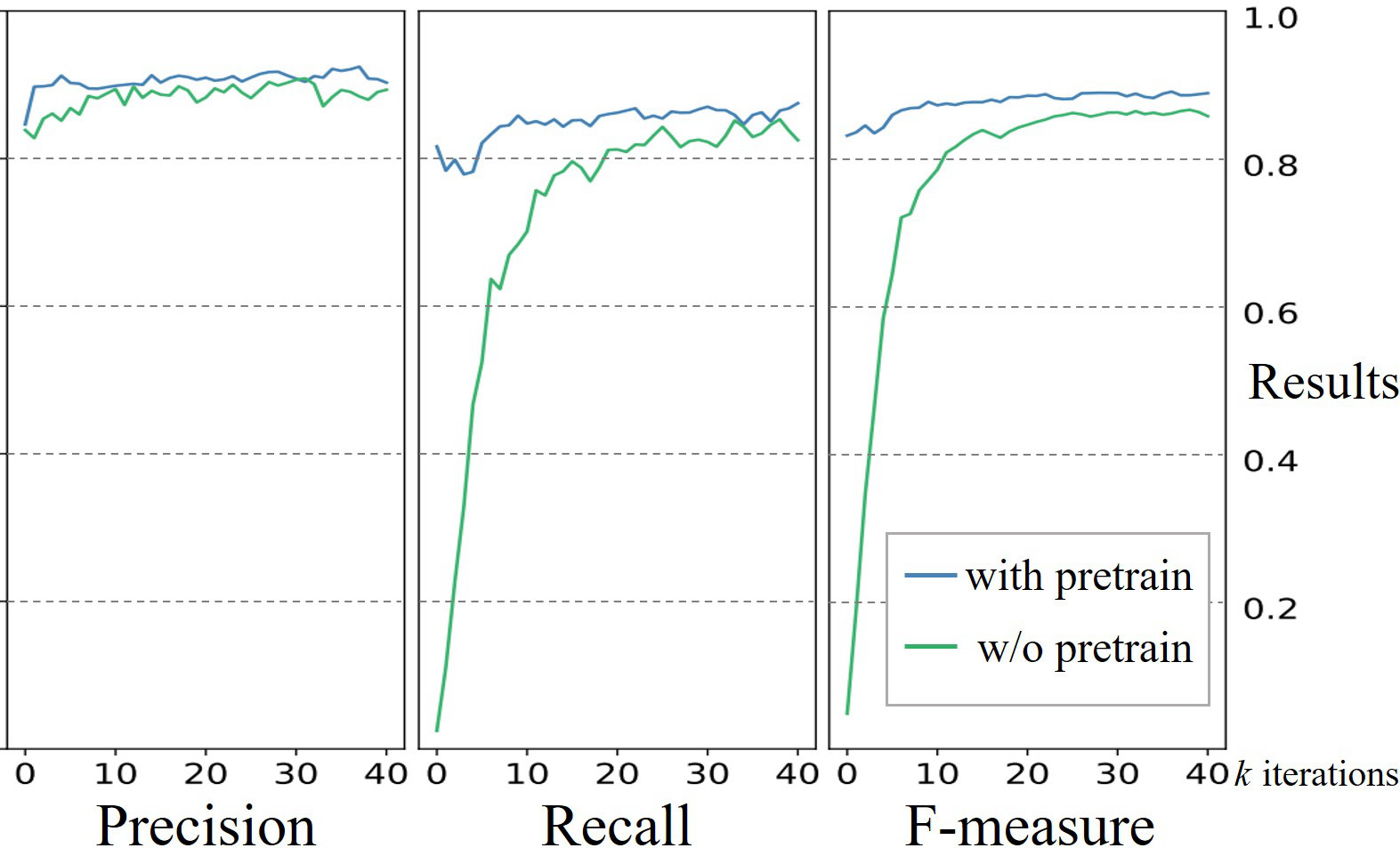}
  \caption{{The convergence curves in model training. The curves start at 500 iterations. We can observe the model with pretraining achieve much higher recall than the without pretraining at the initial iterations.}}
  \label{fig:convergence}
\end{figure}

\begin{table*}[t]
  \centering
  {
  \caption{{Comparison with previous text detection methods on the curved text datasets (Total-Text and CTW1500). ``Ext'' means the external dataset used in the training stage. ``ST'' and ``MLT'' denote the synthetic and ICDAR-MLT dataset. Bold and underlined texts denote the top result and the runner-up, respectively.}}
  \label{tab:benchmarks}
  \footnotesize
  \begin{tabular}{l|c|ccc|c|ccc}
  \hline
  \multirow{2}{*}{Method} & \multirow{2}{*}{Ext} & \multicolumn{3}{c|}{Total-Text} & \multirow{2}{*}{Ext} & \multicolumn{3}{c}{CTW1500}\\ 
  ~ & ~ & P & R & F  & ~ & P & R & F \\ 
  \hline
  TextSnake~\cite{long2018textsnake} & ST & 82.7 & 74.5 & 78.4 & ST & 67.9 & 85.3 & 75.6 \\
  TextField~\cite{xu2018textfield} & ST+MLT & 81.2 & 79.9 & 80.6 & ST+MLT & 83.0 & 79.8 & 81.4  \\
  CRAFT~\cite{baek2019character} &ST& 87.6 & 79.9 & 83.6 & ST & 86.0 & 81.1 & 83.5 \\
  DB~\cite{liao2020real} & ST & 87.1 & 82.5 & 84.7 & ST & 86.9 & 80.2 & 83.4 \\
  ContourNet~\cite{wang2020contournet} & -  & 86.9 & 83.9 & 85.4 & - & 83.7 & 84.1 & 83.9 \\
  DRRG~\cite{zhang2020deep} &ST & 86.5 & 84.9 & 85.7 & ST & 85.9 & 83.0 & 84.5 \\
  MAYOR~\cite{qin2021mask} & MLT & 90.7 & 87.1 & 88.9 & ST & 88.7 & 83.6 & 86.1 \\
  TextFuseNet~\cite{ye2020textfusenet} & ST & 87.5 & 83.2 & 85.3 & ST & 85.8 &85.0 & 85.4  \\
  PCR~\cite{dai2021progressive} & MLT & 88.5 & 82.0 & 85.2 & MLT & 87.2 & 82.3 & 84.7  \\
  TextPMs~\cite{zhang2022arbitrary} & ST & 90.0 & \underline{87.7} & 88.8 & ST & 87.8 & 83.8 & 85.8  \\
  \hline
  LOMO-MS~\cite{zhang2019look} & ST & 87.6 & 79.3 & 83.3 & ST & 85.7 & 76.5 & 80.8 \\
  TextMountain-MS~\cite{zhu2021textmountain} & MLT & - & - & - & MLT & 82.9 & 83.4 & 83.2 \\
  TPSNet-MS~\cite{wang2022tpsnet} & ST & 90.2 & 86.8 & 88.5 & ST & 88.7 & 86.3 & 87.5 \\
  PSENet~\cite{wang2019shape} & MLT & 84.0 & 78.0 & 80.9 & MLT & 84.8 & 79.7 & 82.2\\
  PAN~\cite{wang2019efficient} & ST & 89.3 & 81.0 & 85.0 & ST & 86.4 & 81.2 & 83.7 \\
  TextBPN~\cite{zhang2021adaptive} & MLT & 90.7 & 85.2 & 87.9 & MLT & 86.5 & 83.6 & 85.0 \\
  MSR~\cite{xue2019msr} & ST & 83.8 & 74.8 & 79.0 & ST & 85.0 & 78.3 & 81.5 \\
  \hline
  Tang \etal\cite{tang2022few} & ST & 90.7 & 85.7 & 88.1 & ST & 88.1 & 82.4 & 85.2 \\
  DPText-DETR~\cite{ye2022dptext} &ST+MLT& 91.8 & 86.4 & \underline{89.0} &ST+MLT& \underline{91.7} & \underline{86.2} & \bf{88.8}\\
  \hline
  ABCNetV2~\cite{liu2021abcnet} &ST+MLT& 90.2 & 84.1 & 87.0 &ST+MLT& 85.6 & 83.8 & 84.7 \\
  TESTR-polygon~\cite{zhang2022text} &ST+MLT& \bf{92.8} & 83.7 & 88.0 &ST+MLT & \bf{92.0} & 82.6 & 87.1 \\
  \hline
  \textbf{\system} & ST & \underline{91.9} & \bf{88.3} & \bf{90.1} & ST & 89.7 & \bf{87.9} & \bf{88.8} \\
  \hline
  \end{tabular}
  }
\end{table*}

\vspace{0.03in}
\noindent\textbf{Text Transformer components.} 
We evaluate a few parameters and an alternative implementation for generating \system. The first is the design of the projection module. We consider both the linear projection and a stack of convolution/residual blocks. As listed in Table~\ref{tab:components}(a), we found that using residual blocks significantly boosts the recall for text detection. We also evaluate the number of residual blocks for the projection module. The performance gain is observed when the number increases from 2 to 3. Adding more residual blocks does not improve recall or precision, probably due to the features in each token being too coarse to capture text information for small text regions.

The effect of the Transformer encoder in multi-scale representation is examined in Table~\ref{tab:components}(b). By replacing the Transformer encoder with the FPN-like convolutional encoder and the same projection and decoders, we obtain a precision of 91.0 and recall of 85.6, which are not as good as the Transformer encoder in our framework. 

Our last alternative experiment evaluates the number of the scale-wise decoder in the decoder module (Table~\ref{tab:components}(c)). Using only 3 decoders, we can already get an F-measure of 89.0, while the F-measure without any decoder is 75.0. Adding more decoders boosts the results, and the overall performance tends to be saturated.

\vspace{0.03in}
\noindent\textbf{Model training.} \fig~\ref{fig:convergence} plots the convergence curves to train our network. The model pretraining is important for a stable network, as the image number of the Total-Text training set is limited. 
Directly applying the pre-training model on Total-Text achieves a precision of 76.7 and a recall of 48.8. With only 500 iterations of training with Total-Text training images, the precision and recall reach 84.6 and 81.6 respectively. 
The network converges after 30$k$-40$k$ iterations; training with more iterations does not improve the results.

\subsection{Comparison with State-of-the-Arts}

\begin{table}[t]
  \centering
  {
  \caption{{Results on ICDAR2019-ART.}}
  \label{tab:benchmarks_mlt}
  \footnotesize
  \begin{tabular}{l|c|ccc}
  \hline
  Method & Ext & P & R & F \\ 
  \hline
  CRAFT~\cite{baek2019character} & ST & 77.2 & 68.9 & 72.9 \\
  TextFuseNet~\cite{ye2020textfusenet} & ST & 82.6 & 69.4 & 75.4 \\
  PCR~\cite{dai2021progressive} & MLT & 84.0 & 66.1 & 74.0 \\
  TPSNet-MS~\cite{wang2022tpsnet} & ST & \underline{84.3} & 73.3 & \underline{78.4}\\
  DPText-DETR~\cite{ye2022dptext} & ST+MLT & 83.0 & \underline{73.7} & 78.1\\
  \hline
  \textbf{\system} & ST & \bf{85.3} & \bf{77.5} & \bf{81.2} \\
  \hline
  \end{tabular}
  }
\end{table}

\noindent\textbf{Curved text detection.} We compare the proposed ATTR with several baselines and state-of-the-art approaches. Among them, the first group contains several well-known scene text detectors, \eg, TextSnake~\cite{long2018textsnake}, CRAFT~\cite{baek2019character}, and DB~\cite{liao2020real}. We also compare our approach with previous multi-scale aggregation methods, including the late fusion of multi-scale image pyramid (\eg, LOMO-MS~\cite{zhang2019look}, TextMountain-MS~\cite{zhu2021textmountain} and TPSNet-MS~\cite{wang2022tpsnet}), feature-level scale aggregation (\eg, PSENet~\cite{wang2019shape} and TextBPN~\cite{zhang2021adaptive}), and multi-scale features network (MSR~\cite{xue2019msr}). As our framework is based on Vision Transformer, recent Transformer-based text detection methods are evaluated, including~\cite{tang2022few} and DPText-DETR~\cite{ye2022dptext}. The last group is the text spotting methods that incorporate the text semantic, \ie, ABCNetV2~\cite{liu2021abcnet} and TESTR-polygon~\cite{zhang2022text}.

\begin{table*}[h]
  \centering
  \caption{{Comparison with previous methods on the multi-oriented text datasets (ICDAR 2015 and MSRA-TD500).}}
  \footnotesize
  \label{tab:multi-oriented}
  {
    \begin{tabular}{l|c|ccc|c|ccc}
      \hline
      \multirow{2}{*}{Method} & \multirow{2}{*}{Ext} & \multicolumn{3}{c|}{(a) ICDAR 2015} & \multirow{2}{*}{Ext} & \multicolumn{3}{c}{(b) MSRA-TD500}\\ 
      ~ & ~ & P & R & F  & ~ & P & R & F \\ 
      \hline
    TextSnake~\cite{long2018textsnake} & ST & 84.9  & 80.4  & 82.6 & ST & 83.2 & 73.9 & 78.3 \\
    TextField~\cite{xu2018textfield}  & MLT & 84.3 & 83.9 & 84.1 & ST+MLT & 87.4 & 75.9 & 81.3 \\
    CRAFT~\cite{baek2019character}  & ST & 89.8 & 84.3 & 86.9 & ST & 88.2 & 78.2 & 82.9 \\
    DB~\cite{liao2020real}  & ST & \underline{91.8} & 83.2 & 87.3 & ST & 91.5 & 79.2 & 84.9 \\
    ContourNet~\cite{wang2020contournet} & - & 87.6 & 86.1 & 86.9 & - & - & - & -\\
    DRRG~\cite{zhang2020deep}  & ST & 88.5 & 84.7 & 86.6 & ST & 88.1 & 82.3 & 85.1 \\
    MAYOR~\cite{qin2021mask}  & - &  90.5 & 85.2 & 87.8 & - & \bf{91.7} & 85.2 & 88.3 \\
    PCR~\cite{dai2021progressive} & - & - & - & - & - & 77.8 & \underline{87.6} & 82.4  \\
    TextPMs~\cite{zhang2022arbitrary}  &  ST & 89.9 & 84.9& 87.4 & ST & 91.0 & {86.9} & \underline{88.9} \\
    \hline
    PSENet~\cite{wang2019shape} & MLT & 86.9 & 84.5 & 85.7 & -& -&-&-\\
    LOMO-MS~\cite{zhang2019look} & ST & 87.8  & \underline{87.6}  & 87.7 & -& -& -&- \\
    Lyu \etal~\cite{lyu2018multi}  & ST &\bf{93.3}  & 79.4 & 85.8& ST & 87.6 & 76.2 & 81.5 \\
    PAN~\cite{wang2019efficient}  & ST & 84.0 & 81.9 & 82.9  & ST & 84.4 & 83.8 & 84.1 \\
    TextBPN~\cite{zhang2021adaptive}  & - & - & - & - & MLT & 86.6 & 84.5 & 85.6 \\
    MSR~\cite{xue2019msr}  & ST & 86.6 & 78.4 & 82.3& ST & 87.4 &76.7 &81.7 \\
    \hline
    Tang \etal~\cite{tang2022few}  & ST & 90.9 & {87.3} & \underline{89.1} & ST & \underline{91.6} & 84.8 & 88.1 \\
    Raisi\etal~\cite{raisi2021transformer}  & ST &89.8  &78.3  & 83.7& ST &90.9 & 83.8 &87.2 \\
    \hline
    ABCNetV2~\cite{liu2021abcnet} &ST+MLT & 90.4 & 86.0 &88.1 & ST+MLT& 89.4 & 81.3 & 85.2 \\
    TESTR-polygon~\cite{zhang2022text} & ST+MLT & 90.3  & \bf{89.7} & \bf{90.0} & - & - & - & - \\
    \hline
    \textbf{\system}  & ST & 90.1 & {87.3} & 88.7 & ST & 91.3 & \bf{90.5} & \bf{90.9} \\
    \hline
    \end{tabular}
  }
\end{table*}

\begin{figure}[t]
  \centering
  \includegraphics[width=\linewidth]{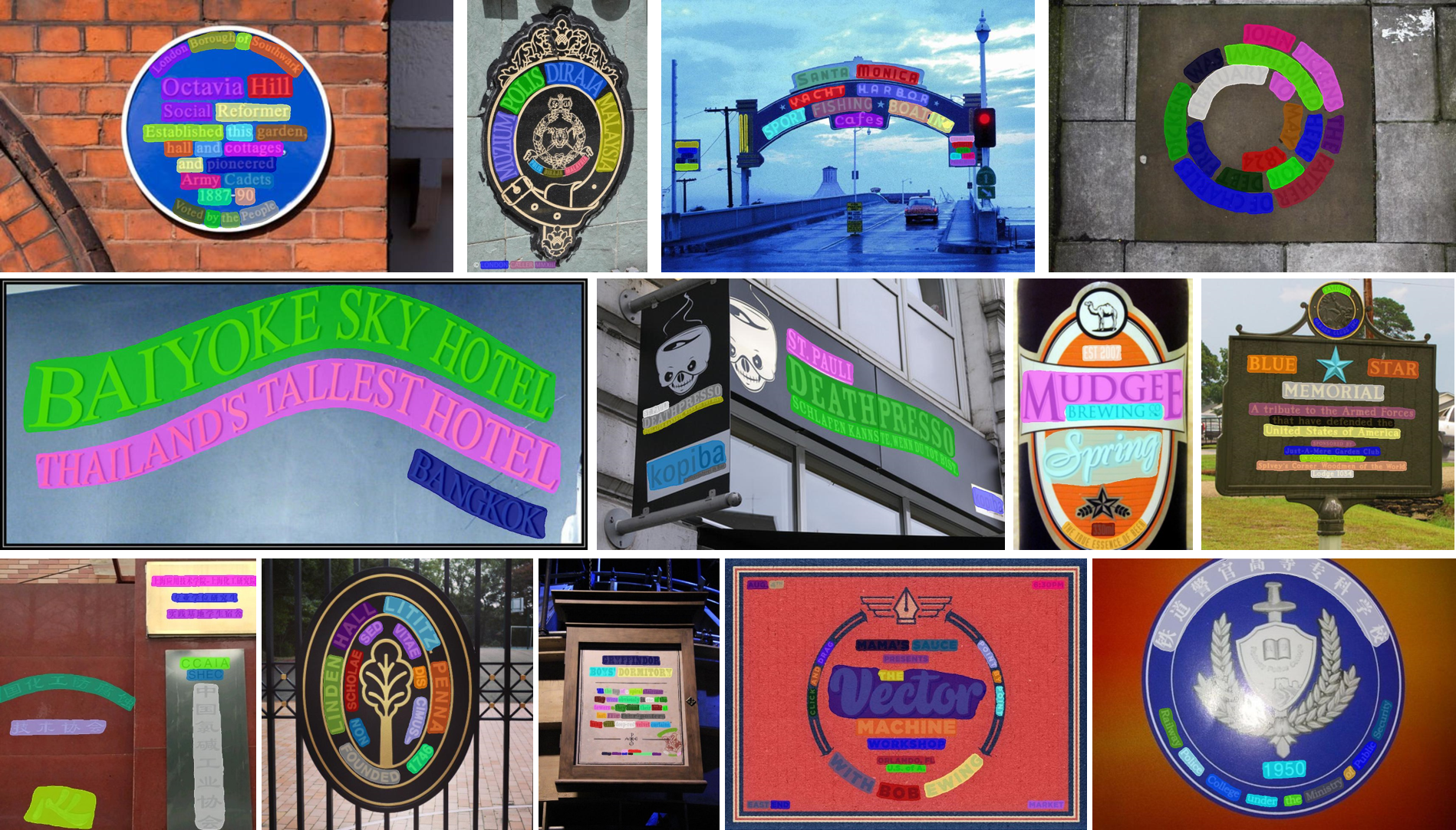}\\
  (a) The curved text datasets.\\
  \vspace{0.03in}
  \includegraphics[width=\linewidth]{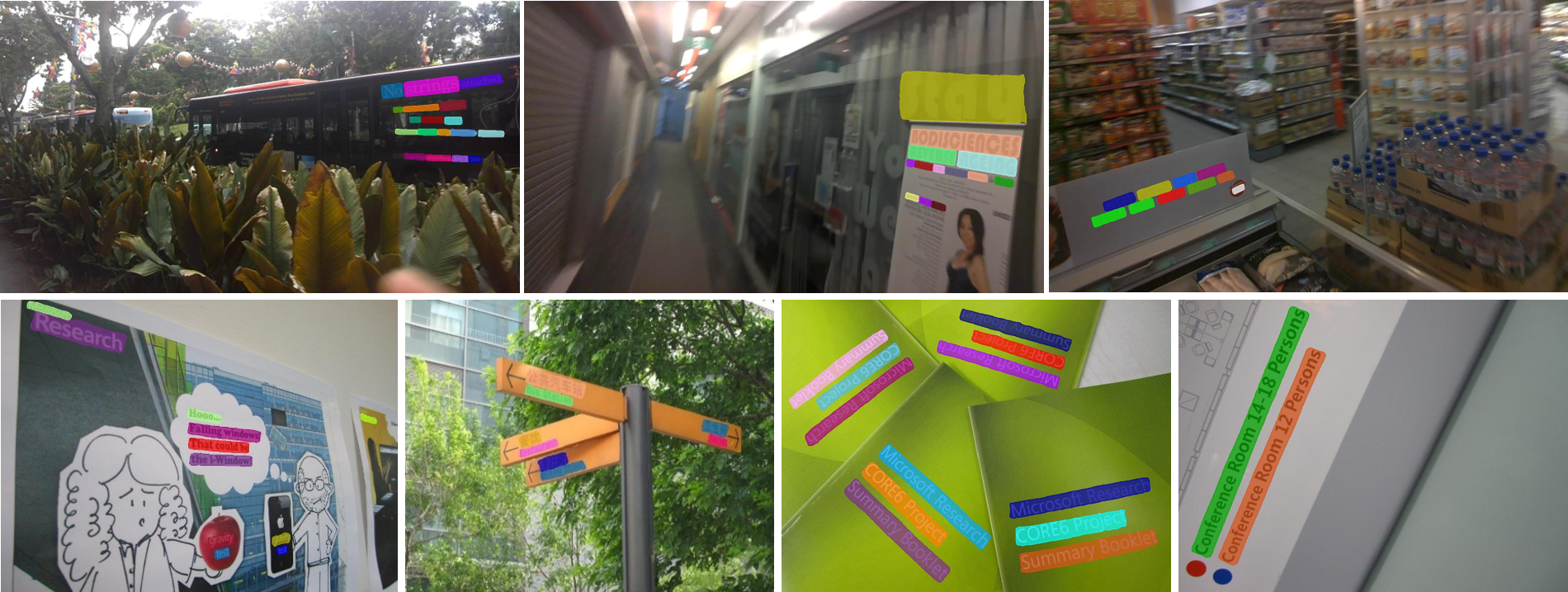}\\
  (b) The multi-oriented text datasets.
  \caption{Result visualization: (a) TotalText, CTW1500, ICDAR2019-ART; (b) ICDAR2015, MSRA-TD500.}
  \label{fig:result}
\end{figure}

Table~\ref{tab:benchmarks} and Table~\ref{tab:benchmarks_mlt} summarize the overall performances of previous methods and ours in the benchmarks. \system shows very competitive performance with the F-measure for all three datasets, which outperforms the previous methods in Total-Text and ICDAR2019-ART. Particularly, the recall for text detection is significantly better than these methods for all three datasets, which confirms the coverage of text instances for the proposed aggregated text Transformer. The precision rate reaches the same magnitude as that of DPText-DETR~\cite{ye2022dptext} and TESTR-polygon~\cite{zhang2022text} in Total-Text, but lower than them in CTW1500, probably due to the difference in the pre-training set. Compared with MSR~\cite{xue2019msr} that also considers multi-scale network and image pyramid as input, the \system performs better in all metrics. 
Moreover, in order to demonstrate the robustness of our model for different languages, we evaluate it on large-scale curve scene text datasets (ICDAR2019-ART). Compared with the state-of-the-art model~\cite{ye2022dptext}, the ATTR improves in terms of precision, recall, and F-measure by 2.3, 3.8, and 3.1, while our model uses only the synthetic dataset for pre-training. The text detection results are shown in {\fig~\ref{fig:result}(a)}.

\vspace{0.03in}
\noindent\textbf{Multi-oriented text detection.} For the images with multi-oriented texts, we use the ICDAR2015 and MSRA-TD500 datasets for evaluation.  
As shown in Table~\ref{tab:multi-oriented}(a), our method is comparable with state-of-the-art methods in the ICDAR2015 dataset.

\begin{table}[t]
For the MSRA-TD500 dataset (Table~\ref{tab:multi-oriented}(b)), the proposed ATTR outperforms the previous method by F-measure of 2.0 on the MSRA-TD500 dataset, which demonstrates that our method is capable of handling both multi-oriented texts and the long texts. Some multi-oriented text detection results are shown in {\fig~\ref{fig:result}(b)}.
  \centering
    \caption{{Text detection results using TIoU measurement metrics. ``T-P'', ``T-R'', ``T-F'' represent TIoU-Precision, TIoU-Recall, and TIoU-F-measure.}}
    \footnotesize
    \label{tab:TIOU}
    \begin{tabular}{l|ccc|ccc}
      \hline
      \multirow{2}{*}{Method} & \multicolumn{3}{c|}{{CTW1500}}  & \multicolumn{3}{c}{{TotalText}} \\
      & T-P & T-R & T-F & T-P & T-R & T-F\\
      \hline
      TextBPN~\cite{zhang2021adaptive} & 69.2 & 61.5 & 65.1 & 68.6 & 58.1 & 63.0 \\
      TPSNet~\cite{wang2022tpsnet} & \underline{72.4} & \underline{62.9} & \underline{67.3} & 68.8 & 61.5 & 65.0\\
      DBNet++~\cite{liao2022real} & 63.8 & 50.8 & 56.5 & \underline{70.5} & 56.5 & 62.7\\
      ABCNetV2~\cite{liu2021abcnet} & 69.2 & 62.0 & 65.4 & 68.6 & \underline{62.6} & \underline{65.5}\\
      \hline
      \textbf{\system} & \bf{75.1} & \bf{67.8} & \bf{71.3} & \bf{72.6} & \bf{64.9} & \bf{68.5}\\
      \hline
      \end{tabular}
\end{table}

\begin{figure}[t]
  \centering
  \setlength{\abovecaptionskip}{5px}
  \subfigbottomskip=-3pt
  \subfigcapskip=-5pt
  \subfigure[ABCNetV2 \cite{liu2021abcnet}]{\includegraphics[width=0.45\linewidth]{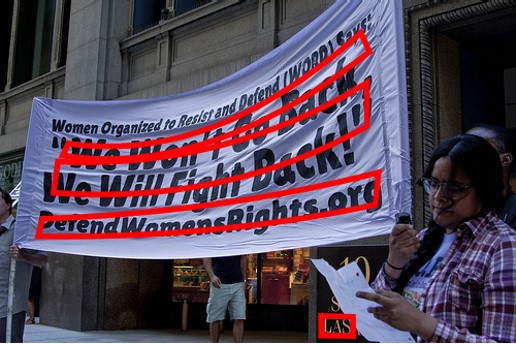}}\hspace{1.5pt}
  \subfigure[DBNet++ \cite{liao2022real}]{\includegraphics[width=0.45\linewidth]{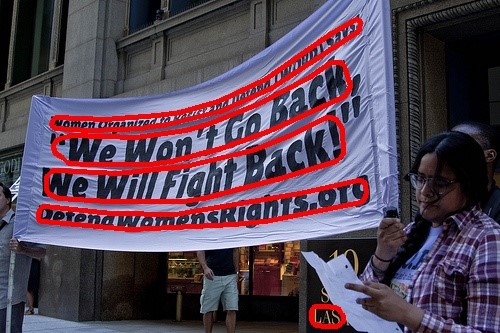}}\vspace{-2pt} \\
  \subfigure[TPSNet \cite{wang2022tpsnet}]{\includegraphics[width=0.45\linewidth]{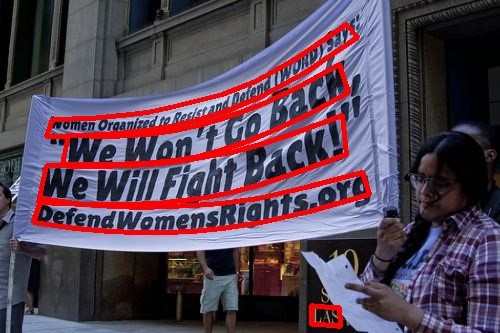}}\hspace{1.5pt}
  \subfigure[ATTR (Ours)]{\includegraphics[width=0.45\linewidth]{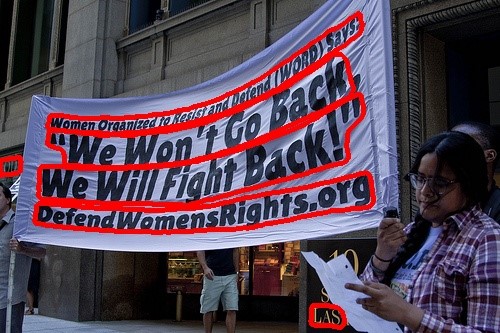}}
  \caption{{Comparison of detection results from different text representations on selected challenging samples in CTW1500. The proposed \system can detect text instances with more complete and compact boundary.}}
  \label{fig:text_represent_comp}
\end{figure}

\subsection{More Results and Discussions}

\vspace{0.03in}
\noindent\textbf{TIoU on curve texts.} Besides the standard protocol, we also evaluate with the TIoU metrics. TIoU~\cite{liu2019tightness} is an evaluation protocol for more accurate detection of the scene texts and has a stricter criterion than the standard metric. 
The TIoU-Recall and TIoU-Precision assess the completeness and compactness of the detection results, respectively, while TIoU-F measures the overall tightness. Here we report the results on the CTW1500 and TotalText datasets, as the completeness and compactness of the curve texts are usually more easily affected by the inaccurate detection. 
We use the official codes and models of previous method~\cite{zhu2021fourier,dai2021progressive,zhang2021adaptive,wang2022tpsnet,liao2022real,liu2021abcnet}, and the TIoU evaluation codes are from \cite{liu2019tightness}.
As presented in Table~\ref{tab:TIOU}, the proposed ATTR is significantly higher than these methods on both datasets. 
Some qualitative comparisons with previous text representations are illustrated in \fig~\ref{fig:text_represent_comp}.
In the CTW1500 dataset, the improvement of our method over the previous method  (TPSNet~\cite{wang2022tpsnet}) is 4.0 (TIoU-F), while the gap under the standard F-measure is 1.6. 
In the TotalText dataset, the improvement of our method over the previous method  (ABCNetV2~\cite{liu2021abcnet}) is 3.0 (TIoU-F), and the gap for F-measure is 1.3.
The substantial performance gains confirm the effectiveness of using ATTR for tighter text representation.

\begin{table}[t]
  \centering
  \caption{Results with different size of text instances.}
  \footnotesize
  \label{tab:large-small}
  {
    \begin{tabular}{l|p{0.1in}<{\centering}p{0.1in}<{\centering}p{0.15in}<{\centering}|p{0.1in}<{\centering}p{0.1in}<{\centering}p{0.15in}<{\centering}|p{0.1in}<{\centering}p{0.1in}<{\centering}p{0.15in}<{\centering}}
      \hline
      \multirow{2}{*}{Method} & \multicolumn{3}{c|}{Large} & \multicolumn{3}{c|}{Medium} & \multicolumn{3}{c}{Small}\\ 
      ~ & P & R & F & P & R & F & P & R & F \\ 
      \hline
      ABCNetV2~\cite{liu2021abcnet} & 95.6 & 87.6 & 91.4 & {\bf 98.3} & 84.6 & 90.9 & \underline{81.0} & 69.8 & 75.0  \\
      DPText-DETR~\cite{ye2022dptext}  & {\bf 97.5} & \underline{89.4} & \underline{93.3} & \underline{97.4} & \underline{91.5} & \underline{94.3} & 78.0 & \underline{75.8} & \underline{76.9} \\
    \hline
    \textbf{\system} & \underline{95.9} & {\bf 92.4} & {\bf 94.1} & 95.8 & {\bf 94.6} & {\bf 95.2} & {\bf 88.0} & {\bf 79.9} & {\bf 83.8} \\
    \hline
    \end{tabular}
  }
\end{table}

\vspace{0.03in}
\noindent
{\textbf{Results on text instances with different scales.} 
We divide the detection text objects based on their proportion in the entire image. When the proportion is larger than 2\% (smaller than 0.2\%), the instance is categorized as ``Large'' (``Small''); otherwise, it is categorized as ``Medium''. The results of both the proposed \system and two state-of-the-art methods are presented in Table \ref{tab:large-small}.
The proposed \system achieves high F-measures among scale categories compared with previous methods, especially for the ``Small'' scale with a margin of 6.9.
Quantitatively speaking, the results of the recall rate perform the best among different algorithmic settings, indicating the robustness over different sizes of text instances.
}

\begin{table}[t]
  \centering
  \footnotesize
  \caption{{Effect of the alternative with previous methods. ``PEM'' (``DM'')  indicates using the projection and encoder module (decoder module) of \system.}}
  \label{tab:alternative}
  {
    \begin{tabular}{c|cc|ccc|c}
      \hline
      Method & PEM & DM & P & R & F & fps\\
      \hline
      \system & \checkmark & \checkmark & 91.9 & 88.3 & 90.1 & 4.6 \\
      \hline
      Mask2Former~\cite{cheng2022masked} &  & \checkmark &  88.4 & 84.8 & 86.5 & 6.5\\
      Detection head of DB~\cite{liao2020real} & \checkmark &  & 88.0 & 84.0 & 86.0 & 4.2\\
      \hline
      DB~\cite{liao2020real} &  &  & 87.1 & 82.5 & 84.7 & 19.3\\ 
      DPText-DETR~\cite{ye2022dptext} &  &  & 91.8 & 86.4 & 89.0 & 6.0\\
        \hline
        \end{tabular}
  }
\end{table}

\vspace{0.03in}
\noindent
{\textbf{Alternative with previous methods.}  
To validate the effectiveness of the components of our methods, we further design the alternative with previous methods (Table~\ref{tab:alternative}).
As our decoder module is inspired by~\cite{cheng2022masked}, we first compare it with a different encoder design. The results show that our framework is more suitable for the scene text detection task.
We also conduct an experiment for a combination of our projection and encoder module with the detection head from DB \cite{liao2020real}.
Compared with the original architecture of DB~\cite{liao2020real}, using our backbone design yields a performance improvement of 1.3 for the F-measure (84.7 to 86.0), while DB has better inference speed.
With the same backbone and encoder module, we observe significant performance gains when employing the \system decoders instead of the detection head of DB.
Notably, the inference speed is slightly faster than that for the detection head of DB, as the post-processing for DB is more complex.}

\vspace{0.03in}
\noindent{\textbf{Inference speed.}} 
{Under the default setting of \system (image pyramid with 3 scales, 3 residual blocks for the projection module, and 9 scale-wise decoders), we get 4.6 fps on the TotalText dataset\footnote{{The inference time for the main modules: 0.016s for the projection module, 0.121s for the encoder and 0.060 for decoder.}}.
We also report fps for different settings of \system in the right part of Table~\ref{tab:components}. Employing linear projection or a simple convolutional layer for the projection module is faster than with the residual blocks, however the performance is lower. Using fewer residual blocks leads to a larger feature map and more tokens, which results in a decrease in the overall speed. The speed of the Transformer encoder in ATTR is slower than the FPN-like structure (Table~\ref{tab:aggregation}), due to the computational cost from multiple image pyramids. For the decoder part, we can observe that adding more decoder layers slightly impacts on inference time.
We also compare the inference time with existing methods. As shown in Table~\ref{tab:alternative}, our method exhibits slightly slower inference speeds compared with the state-of-the-art Transformer-based method (DPText-DETR, fps 6.0), and is slower than the non-transformer-based method (DB, fps 19.3).
}

\vspace{0.03in}
\noindent{\textbf{Limitations.} }
Although the proposed method is capable of scene text images, there are still limitations. First, as listed in Table~\ref{tab:ms} and Table~\ref{tab:alternative}, employing an image pyramid with multiple scales requests more computational resources and significantly increases the inference time, especially when compared with existing feature pyramid-based methods. Second, following the matching strategy from DETR~\cite{carion2020end}, our method might miss the detection of dense texts with extreme number of text instances that exceed the text query number.

\section{Conclusions}
\label{sec:conclusion}
In this paper, we presented the ATTR, a multi-scale aggregated text Transformer for the scene text detection task. Our framework significantly enhances the baseline text detector for representing texts in scene images with a multi-scale structure and the self-attention mechanism. In addition, we represent text instances by individual binary masks, which can separate neighboring text instances better and get tighter text instance boundaries. Experiments over five public datasets show the effectiveness of the proposed framework.

\bibliographystyle{IEEEtran}
\bibliography{arxiv}

\end{document}